# Performance of Large Language Models in Answering Critical Care Medicine Questions

Mahmoud Alwakeel, MD[1], Aditya Nagori, Ph.D[1], An-Kwok Ian Wong, MD, Ph.D[1], Neal Chaisson, MD,[2] Vijay Krishnamoorthy, MD, Ph.D[1], Rishikesan Kamaleswaran, Ph.D[1],

[1]Duke University School of Medicine, Durham, North Carolina, United States
[2]Cleveland Clinic Foundation, Cleveland, Ohio, United States

**Abstract:**
Large Language Models have been tested on medical student-level questions, but their performance in specialized fields like Critical Care Medicine (CCM) is less explored. This study evaluated Meta-Llama 3.1 models (8B and 70B parameters) on 871 CCM questions. Llama3.1:70B outperformed 8B by 30%, with 60% average accuracy. Performance varied across domains, highest in Research (68.4%) and lowest in Renal (47.9%), highlighting the need for broader future work to improve models across various subspecialty domains.

**Introduction:**
The use of Large Language Models (LLMs) to answer medical exam-style questions has gained popularity in recent years.[1,2] However, most studies have focused on exams targeting medical students, which do not fully capture the complexities of subspecialty fields, like Critical Care Medicine (CCM). This study aims to evaluate the performance of LLMs in answering subspecialty CCM board exam-style questions.

**Methods:**
We conducted an experimental study evaluating two Meta-Llama 3.1 models with 8 and 70 billion parameters on 871 questions covering real-world CCM cases. Meta-Llama is a family of LLMs designed to understand and generate human-like language, with varying parameter sizes to balance performance and computational efficiency. The non-public question bank dataset, created for CCM board preparation in the United States, comprised multiple-choice questions with single correct answers and included reasoning explanations. We tested model performance using in-context learning with chain-of-thought reasoning, multi-shot examples (0 to 5 shots), and varied temperatures (0, 0.2, 0.5, 1). The outcome measure was accuracy, defined as the percentage of correct answers provided by the models across 12 topic domains.

**Results**
Llama3.1:70B consistently outperformed Llama3.1:8B in all domains, with the highest accuracy in Research//Ethics (68.4%), Surgery/Trauma (68.3%), Pharmacology/Toxicology (67.1%), and Hematology/Oncology (65.7%). It had the lowest accuracies in Renal (47.9%), Gastrointestinal (53.1%), and Miscellaneous (51.8%). In contrast, Llama3.1:8B had consistently lower accuracies across all domains, ranging from 25% to 35%. (see Figure 1)

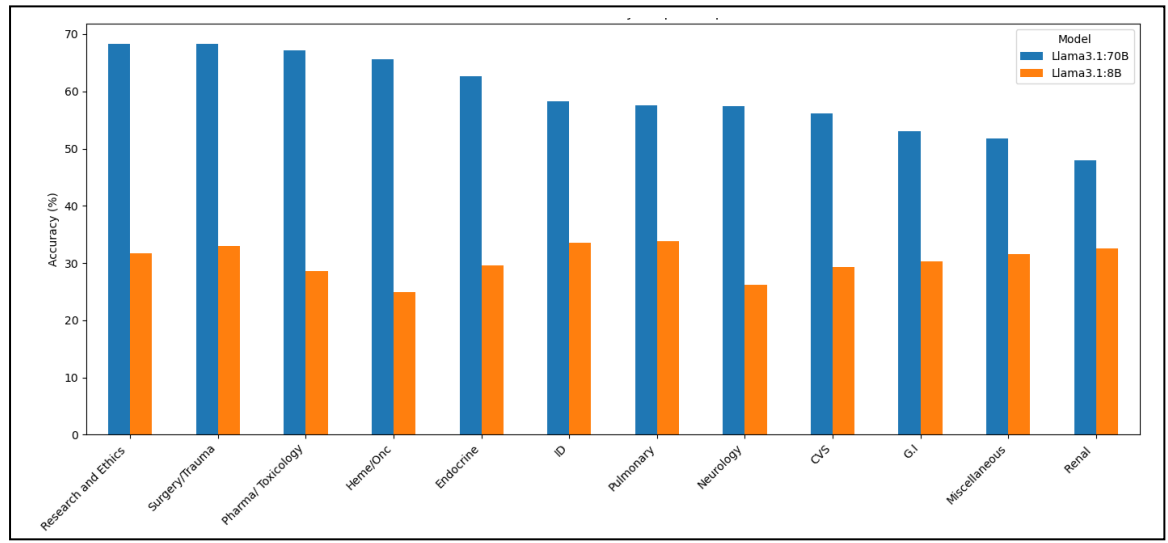

**Figure 1: LLMs Performance in Answering CCM Board Style Questions Over different Domains**

For accuracy by shot number, Llama3.1:70B maintained stable performance at 60% regardless of the number of shots. Llama3.1:8B showed fluctuating accuracy lowest at two shots (26.5%) and highest at 0 shots (36.1%). When assessing accuracy by temperature, Llama3.1:70B remained steady at 59%, while Llama3.1:8B's performance declined slightly as temperature increased, from 31.4% at a temperature of 0.0 to 29.5% at 1.0. (see Figure 2)

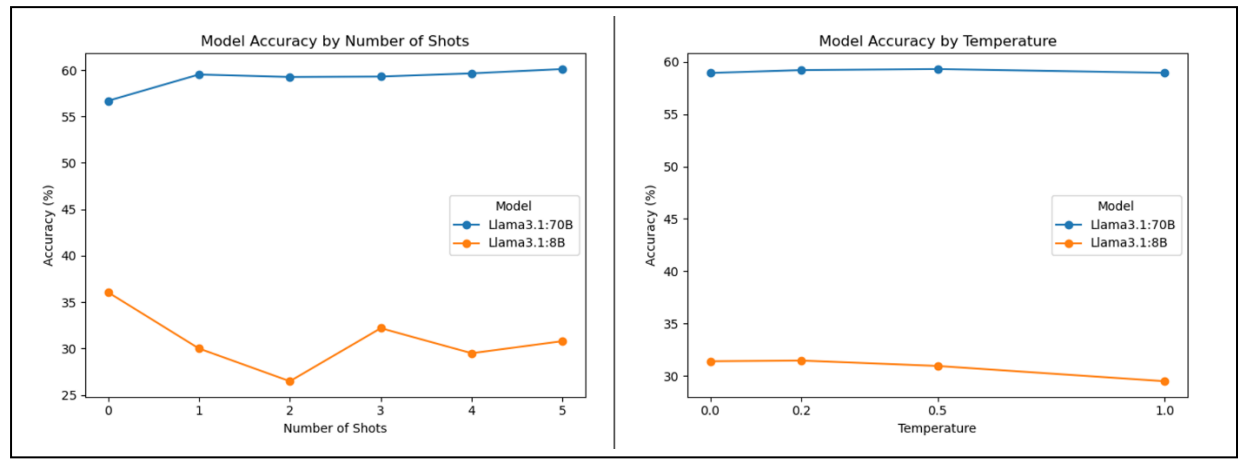

**Figure 2: LLMs Accuracy over different numbers of Shots or Temperatures:**

**Discussion:**
This study is one of the few to evaluate LLM performance in addressing subspecialty board-style exam questions, specifically in CCM. The base model performance varied significantly by clinical domain, highlighting the need for improving those bases models to specific subspeciality domains to improve accuracy and reliability in subspecialties, such as critical care.

Previous LLM studies have focused on medical exams for students, often using widely available materials that may overlap with training data, casting doubt on the reported high performance.[1,2] Katz et al. examined the performance of GPT-3.5 and GPT-4 on Israeli medical board examinations across several specialties.[3] Their study revealed similar findings to ours, showing that higher-parameter models, such as GPT-4, outperformed GPT-3.5, with performance varying by specialty. For instance, GPT-4 ranked in the 75th percentile for psychiatry (95% CI, 66.3–81.0) and near the median for internal medicine (56.65%, 95% CI, 44.0–65.7) and general surgery (44.44%, 95% CI, 38.9–55.6). However, it performed less well in pediatrics (17.4%, 95% CI, 9.5–30.9) and obstetrics/gynecology (23.44%, 95% CI, 14.84–44.53). GPT-3.5 did not pass the examination in any discipline and performed worse than the majority of physicians across all five specialties.

Our study aims to establish a framework for developing a reliable LLM in CCM, beginning with an in-depth analysis of the base model's performance across various CCM clinical domains. From here, we plan to validate our findings using an independent dataset and to broaden our testing to include other medical subspecialties and incorporate simulated testing to determine whether further model training or fine tuning will be necessary.

**Conclusion:**
Our study highlights the variability in LLM performance across critical care domains, emphasizing the need to test these base models across various medical subspecialties. It is crucial to conduct in-depth evaluations within each subspecialty domain before considering real-world applications or guiding future training efforts of LLMs in clinical settings.

**References:**
1   Singhal K, Azizi S, Tu T, *et al.* Large language models encode clinical knowledge. *Nature*. 2023;620:172–80.
2   Lucas MM, Yang J, Pomeroy JK, *et al.* Reasoning with large language models for medical question answering. *J Am Med Inform Assoc*. 2024;31:1964–75.
3   Katz U, Cohen E, Shachar E, *et al.* GPT versus resident physicians — A benchmark based on official board scores. *NEJM AI*. 2024;1. doi: 10.1056/aidbp2300192